\documentclass[nohyperref]{article}

\usepackage{microtype}
\usepackage{graphicx}
\usepackage{subfigure}
\usepackage{booktabs} % for professional tables

\usepackage{hyperref}

\usepackage[accepted]{icml2022}

\usepackage{amsmath}
\usepackage{amssymb}
\usepackage{mathtools}
\usepackage{amsthm}

\usepackage{enumitem}
\usepackage{times}
\usepackage{soul}
\usepackage{url}
\usepackage{float}
\usepackage[utf8]{inputenc}
\usepackage[small]{caption}
\usepackage{graphicx}
\usepackage{amsmath}
\usepackage{physics}
\usepackage{amsthm}
\usepackage{amssymb}
\usepackage{multirow}
\usepackage{booktabs}
\usepackage{algorithm}
\usepackage{algorithmic}
\usepackage[section]{placeins}
\usepackage{stfloats}
\usepackage{makecell}

\usepackage[capitalize,noabbrev]{cleveref}

\theoremstyle{plain}

\theoremstyle{definition}

\theoremstyle{remark}

\usepackage[textsize=tiny]{todonotes}

\icmltitlerunning{Generalizing Multimodal Pre-training into Multilingual via \mbox{Language Acquisition}}

\begin{document}

\twocolumn[
\icmltitle{Generalizing Multimodal Pre-training into Multilingual via \mbox{Language Acquisition}}

% It is OKAY to include author information, even for blind
% submissions: the style file will automatically remove it for you
% unless you've provided the [accepted] option to the icml2022
% package.

% List of affiliations: The first argument should be a (short)
% identifier you will use later to specify author affiliations
% Academic affiliations should list Department, University, City, Region, Country
% Industry affiliations should list Company, City, Region, Country

% You can specify symbols, otherwise they are numbered in order.
% Ideally, you should not use this facility. Affiliations will be numbered
% in order of appearance and this is the preferred way.
\icmlsetsymbol{equal}{*}

\begin{icmlauthorlist}
\icmlauthor{Liang Zhang}{ruc}
\icmlauthor{Anwen Hu}{ruc}
\icmlauthor{Qin Jin}{ruc}
\end{icmlauthorlist}

\icmlaffiliation{ruc}{School of Information, Renmin University of China, Beijing, China}

\icmlcorrespondingauthor{Liang Zhang}{zhangliang00@ruc.edu.cn}
% \icmlcorrespondingauthor{Qin Jin}{qjin@ruc.edu.cn}

% You may provide any keywords that you
% find helpful for describing your paper; these are used to populate
% the "keywords" metadata in the PDF but will not be shown in the document
\icmlkeywords{Machine Learning, ICML}

\vskip 0.3in
]

% this must go after the closing bracket ] following \twocolumn[ ...

% This command actually creates the footnote in the first column
% listing the affiliations and the copyright notice.
% The command takes one argument, which is text to display at the start of the footnote.
% The \icmlEqualContribution command is standard text for equal contribution.
% Remove it (just {}) if you do not need this facility.

% \printAffiliationsAndNotice{}  % leave blank if no need to mention equal contribution
% \printAffiliationsAndNotice{\icmlEqualContribution} % otherwise use the standard text.

\begin{abstract}
English-based Vision-Language Pre-training (VLP) has achieved great success in various downstream tasks. Some efforts have been taken to generalize this success to non-English languages through Multilingual Vision-Language Pre-training (M-VLP). However, due to the large number of languages, M-VLP models often require huge computing resources and cannot be flexibly extended to new languages. In this work, we propose a \textbf{M}ulti\textbf{L}ingual \textbf{A}cquisition (MLA) framework that can easily generalize a monolingual Vision-Language Pre-training model into multilingual. Specifically, we design a lightweight language acquisition encoder based on state-of-the-art monolingual VLP models. We further propose a two-stage training strategy to optimize the language acquisition encoder, namely the Native Language Transfer stage and the Language Exposure stage. With much less multilingual training data and computing resources, our model achieves state-of-the-art performance on multilingual image-text and video-text retrieval benchmarks.
\end{abstract}

\section{Introduction}

We are living in a multimodal and multilingual world w. The information we receive in our daily lives may come from different modalities and languages. Therefore, building multimodal and multilingual models to effectively understand such information has attracted much research attention \cite{gella2017image, wehrmann2019language, kim2020mule, burns2020learning}. Recently, Multilingual Vision-Language Pre-training (M-VLP) achieves convincing performance in various cross-lingual cross-modal tasks such as multilingual image-text retrieval \cite{m3p,Zhou_2021_CVPR,ccpmr,mmp,mural} and multimodal machine translation \cite{upoc2}. 
\begin{figure}[htbp]
    \centering
    \setlength{\abovedisplayskip}{1pt} 
    \setlength{\belowdisplayskip}{1pt} 
    \includegraphics[width=\linewidth]{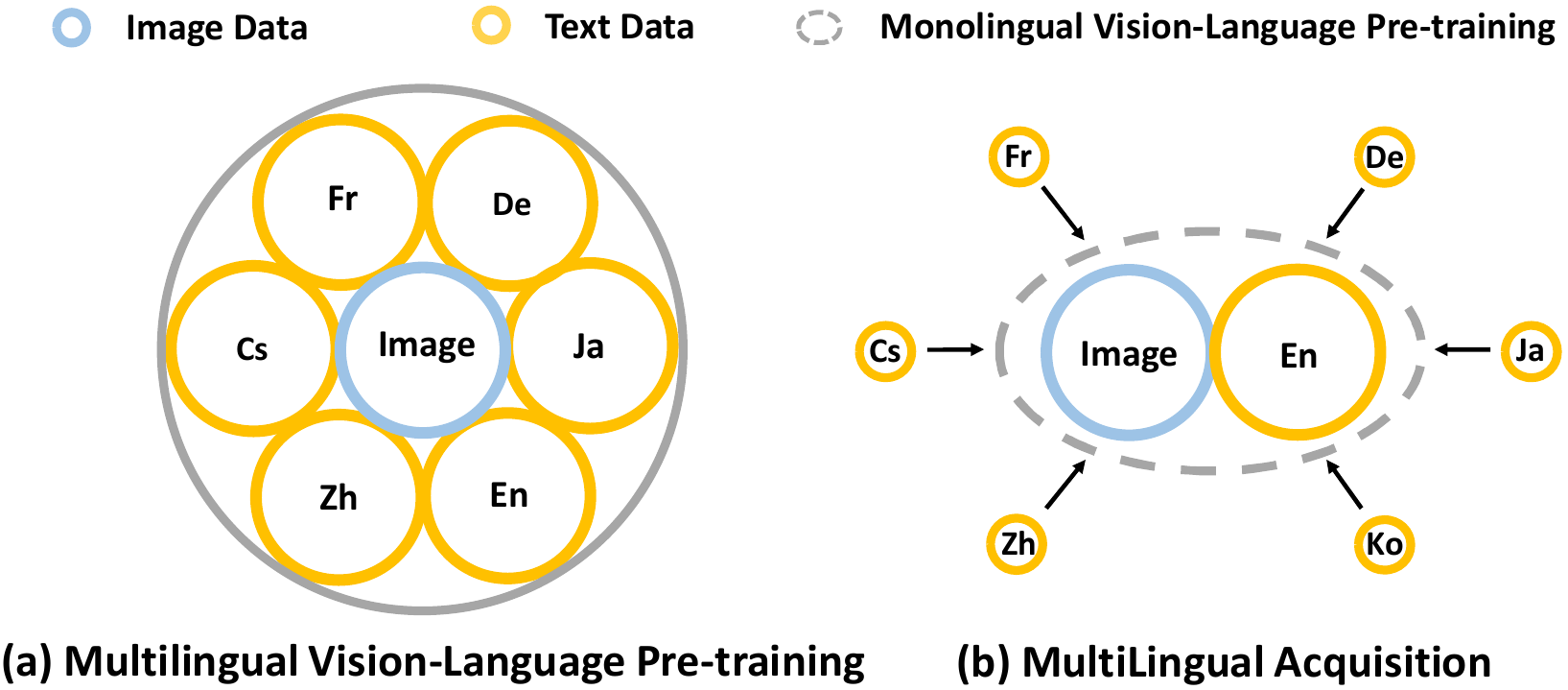}
    \caption{
    Comparison of data usage between M-VLP and MLA. The size of a circle reflects the amount of training data. M-VLPs learn on vision-language data from multiple languages simultaneously. Instead, MLA generalize monolingual VLP into multilingual on much less training data.
    %Topology comparison between Multilingual Vision-Langauge Pre-training (M-VLP) and MutiLingual Acquisition (MLA). The size of a circle reflects the amount of data. Instead of pre-training from scratch, MLA generalizes vision-language pre-training into multilingual through language acquirers. 
    }
    \label{fig: topology}
    \vspace{-12pt}
\end{figure}
As shown in Figure \ref{fig: topology}(a), M-VLP models handle multiple languages and modalities simultaneously during pre-training. Despite their successes, M-VLP models suffer from two problems. First, pre-training on vision and multilingual data consumes huge computing resources. For example, the state-of-the-art M-VLP model MURAL \cite{mural} is pre-trained on 128 Cloud TPUv3 for four days. It could support multimodal tasks on 100+ languages. However, considering there are 6,900+ languages worldwide \cite{Zhou_2021_CVPR}, building such a single model to handle all languages will be highly expensive.
Second, M-VLP models cannot be flexibly extended to new languages. Additional training is required for M-VLP models to achieve satisfactory performance on a new language. However, this training process will cause performance degeneration of M-VLP models on the original languages due to the limited model capacity. For example, the limited model capacity even results in M-VLP models performing worse than their monolingual counterparts on English \cite{m3p, Zhou_2021_CVPR}.

To build multimodal and multilingual models with low-cost and high-flexibility, we refer to our human learning habits when acquiring new languages. We humans normally learn our native language during childhood and practice it through interactions with the multimodal living environments.  
% This learning period is similar to Vision-Language Pre-training (VLP) which is based on a single language.
When learning a new language, we humans initially tend to align it with the native language, as we can easily map words in the native language to real-world objects and concepts. After having a certain language foundation, we could further master it by interacting with the environment directly using the new language. This is known as the language exposure \cite{castello2015first}. The whole learning process rarely degrades our native language capability. %After the whole learning process, out native language ability is rarely affected. 

Inspired by this, we propose a new framework, \textbf{M}ulti\textbf{L}ingual \textbf{A}cquisition (MLA), which constructs multimodal and multilingual models based on monolingual VLPs. The topology of the MLA-based multimodal and multilingual model is illustrated in Figure \ref{fig: topology}(b). Unlike M-VLPs, which handle data from multiple languages and modalities in a single model, MLA generalizes monolingual VLPs into multilingual using much less training data through a language acquisition encoder. The language acquisition encoder is realized by inserting our proposed lightweight language acquirers into the pre-trained monolingual encoder of the VLP model.
%implemented as the pre-trained monolingual encoder of the VLP model inserted with lightweight language acquirers.
During training, original parameters in the pre-trained monolingual encoder are fixed, only multi-lingual embeddings and language acquirers for each new language are optimized. Following the human learning habits, we propose a two-stage training strategy to train the language acquisition encoder. In the Native Language Transfer (NLT) stage, the model is optimized to establish the correspondence between the new languages with the native language. In the Language Exposure (LE) stage, the model is optimized to build cross-modal alignment between new languages and images. We apply our proposed MLA to the monolingual VLP model CLIP \cite{clip} and achieve state-of-the-art results on both multilingual image-text and video-text retrieval benchmarks with much less training data and computing resources. Ablation studies demonstrate the effectiveness of our training strategy. Owing to the independence merit of the language acquirers, the MLA-based models can be easily extended to new languages without compromising the performance of their original languages. 
The main contributions of our work are as follows:
\vspace{-6pt}
\begin{itemize}[leftmargin=*]
\setlength{\itemsep}{0pt}
\setlength{\parsep}{0pt} 
\setlength{\parskip}{0pt} 
    \item We propose a lightweight MultiLingual Acquisition (MLA) framework that can easily generalize monolingual VLPs into multilingual. %The MLA-based models can be flexibly extended to new languages.
    \item We propose a two-stage training strategy to optimize the MLA-based models inspired by the language learning habits of humans. Ablation studies prove the effectiveness of the strategy.
    \item We apply MLA to the monolingual VLP model CLIP and achieve the new state-of-the-art results on both multilingual image-text and video-text retrieval benchmarks with much less training data and parameters.
\end{itemize}

\section{Related Work}
%\paragraph
\noindent\textbf{Vision-Language Pre-training:}
There are increasing interest in building Vision-Language Pre-training (VLP) models. From the perspective of how to interact between vision and language modalities, existing models can be divided into two categories: single-stream and dual-stream models. The single-stream models perform interaction on image and text directly with a cross-modal transformer \cite{uniter, oscar, vilt}. In contrast, the dual-stream models encode image and text with two independent encoders and optimize via simple objectives like image-text contrastive learning \cite{clip, align, florence}. Compared with the single-stream models, the dual-stream models are more efficient to utilize noisy image-text data harvested from the web \cite{wenlan}, and thus achieve better performance and transferability across downstream tasks. Meanwhile, the dual-stream models are more flexible for extension. Since the dual-stream models process images and text through independent encoders, we can fix the vision encoders and focus on extending the text encoders to support new languages. Therefore, we focus on generalizing dual-stream VLPs into multilingual in this work.

%\paragraph
\noindent\textbf{Multilingual Vision-Language Pre-training:}
% To build multilingual multimodal models, multilingual multimodal pre-training methods have been explored in existing works. 
To achieve both multilingual and multimodal capability, many works try to learn the relationship between multiple languages and modalities simultaneously through pre-training. M$^3$P \cite{m3p} introduces the multimodal code-switched training method to enhance multilingual transferability. UC$^2$ \cite{Zhou_2021_CVPR} augments the English image-text data to other languages through machine translation and proposes MRTM and VTLM objectives to encourage fine-grained alignment between images and multiple languages. More recently, MURAL \cite{mural} adopts the dual-stream structure. It is pre-trained with image-text and text-text contrastive objectives on multilingual image-text pairs and translation pairs. M-VLP models significantly outperform previous non-pretraining models \cite{gella2017image, wehrmann2019language, kim2020mule, burns2020learning} on multilingual image-text retrieval. Despite their success, these models typically consume huge computing resources and large-scale multilingual training data. Moreover, they fail to take full advantage of the cross-modal knowledge learnt in monolingual VLP, and building cross-modal cross-lingual representations from scratch can be very hard. In contrast, our MLA framework aims to generalize VLP models into multilingual and it builds multimodal and multilingual models with much less data and computing cost. 
%\paragraph{

\noindent\textbf{Multilingual Extension:}
Some works explore making pre-trained monolingual language models multilingual. Reimers et al. extend sentence embeddings from monolingual to multilingual by Multilingual Knowledge Distillation (MKD)  \cite{mse-kd}. Given translation pairs, MKD optimizes the multilingual student model to produce similar sentence embeddings with the monolingual teacher model. Artetxe et al. extend monolingual models by training additional word embeddings \cite{artetxe-etal-2020-cross}. {MAD-X} \cite{mad-x} extends multilingual pre-training models to support low-resource languages through adapters \cite{adapter}. By extending state-of-the-art pre-trained language models, these works have achieved impressive results in NLP tasks such as bitext retrieval \cite{mse-kd}, cross-lingual QA and NER \cite{mad-x, mse-kd}. However, few works focus on making VLP models multilingual. 
Work in \cite{xgqa} is the first to extend single-stream VLP model OSCAR \cite{oscar}. It adopts a similar strategy with MAD-X \cite{mad-x} that trains language adapters with Masked Language Modeling (MLM) for each language. During inference, it replaces the English language adapters with the target language adapters to achieve zero-shot cross-lingual transfer. However, it generalizes poorly on other languages since the MLM-based training strategy can only implicitly establish the correspondence between other languages and English, let alone vision correspondences. In contrast, MLA directly builds the connection of other languages with English and then with vision in the two-stage training strategy. 
Therefore, MLA achieves comparable results on non-English languages as on English in downstream tasks.

\begin{figure*}
    \centering
    \includegraphics[width=\linewidth]{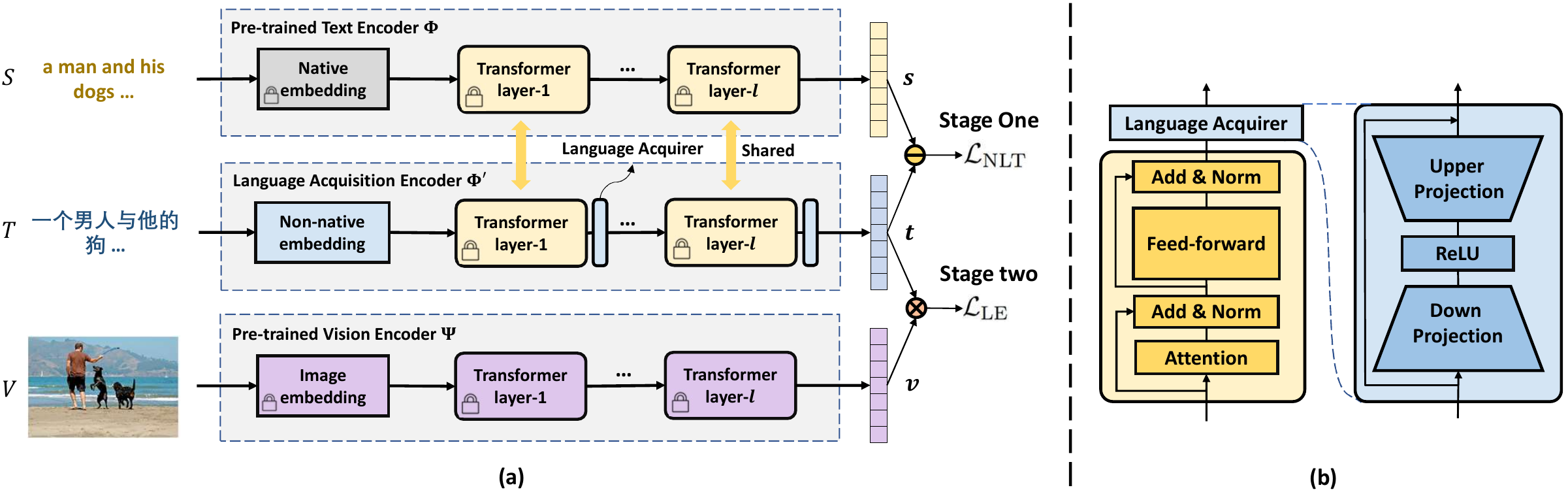}
    \caption{Model illustration: (a) The overview of MLA framework. (b) The structure of a language acquirer}
    \label{fig: architecture}
\end{figure*}

\section{Method}
The MultiLingual Acquisition (MLA) framework is proposed to empower a dual-stream monolingual VLP model with multilingual capability. We define the \textit{native language} of a VLP as its pre-training language. In this paper, we choose CLIP-ViT-B \cite{clip} as the VLP model. It is pre-trained with 400M image-text pairs in English \cite{clip}. Note that MLA can also be applied to VLP models with different native languages. % such as WenLan \cite{wenlan}.

Since the state-of-the-art VLP models can project vision and native language into a shared multimodal space, we design a language acquisition encoder to process non-native languages. 
We then simulate the learning habits of human beings and propose a two-stage training strategy to optimize the language acquisition encoder. We first introduce the architecture of the MLA framework in Sec.\ref{subsec: architecture}. Then, we describe our training strategy in Sec.\ref{subsec: training_strategy}. 

\subsection{Architecture}
\label{subsec: architecture}
Figure \ref{fig: architecture}(a) illustrates the overview of the MLA framework, which consists of three modules: the pre-trained text encoder, the pre-trained vision encoder, and the language acquisition encoder. 

\noindent\textbf{Pre-trained Text Encoder.}
Given a sentence $S$ in the native language, the corresponding sentence representation $\boldsymbol s = \Phi(S; \theta_{\Phi})$ is generated through the pre-trained text encoder $\Phi$. To preserve the cross-model knowledge of VLP, $\theta_\Phi$ is keep fixed during training. As shown in the top part of Figure \ref{fig: architecture}(a), the pre-trained text encoder contains a native embedding block and $l$ transformer layers \cite{transformer}. The native embedding block first tokenizes $S$ with byte pair encoding (BPE) \cite{bpe}. Then, it converts words into embeddings $E_S$ = $[e_{0=\texttt{[SOS]}}, e_1, \dots, e_{M=\texttt{[EOS]}}]$. \texttt{[SOS]} and \texttt{[EOS]} are special tokens denoting the boundary of $S$. The word embeddings are then passed through the transformer layers:
\begin{align}
H^0 &= [e_{0=\texttt{[SOS]}}, e_1, \dots, e_{M=\texttt{[EOS]}}] + E_{pos} \\
\label{eq: native_transformer}
H^i &= \texttt{TransformerLayer}(H^{i-1}; \theta_{\Phi}^i) 
\end{align}
where $H^i=[h^i_0, \dots, h^i_M]$ is the hidden state of the layer $i$. $\theta_\Phi^i$ denotes the parameters of the layer $i$. $E_{pos}$ is the positional encoding. Note that the causal self-attention mask is used in the transformer layers \cite{clip}. The last hidden state of the \texttt{[EOS]} token is chosen to generate the sentence representation:
\begin{align}
\label{eq: native_representation}
 \boldsymbol s = W_a h^l_M
\end{align}
where $\boldsymbol s$ is the sentence representation of $S$, and $W_a$ denotes a linear projection. 

\noindent\textbf{Pre-trained Vision Encoder.} 
We extract the representation $\boldsymbol v=\Psi(V; \theta_{\Psi})$ of an image $V$ with the pre-trained vision encoder $\Psi$. Similar with the pre-trained text encoder, $\theta_\Psi$ is also frozen. The pre-trained vision encoder is implemented as a Vision Transformer \cite{vit}. As shown in the bottom part of Figure \ref{fig: architecture}(a), it consists of a image embedding block and $l$ transformer layers. Given an image $V$, the image embedding block first divides $V$ into patches $V'$ = $[v_1', \dots, v_N']$ following \cite{vit}. Then, they are linearly projected into patch embeddings $E_p=[e_\texttt{[CLASS]}, W_p v_1', \dots, W_p v_N']$, 
where $e_\texttt{[CLASS]}$ is a special embedding for the whole image and $W_p$ is the linear projection. The patch embeddings are then fed into transformer layers:
\begin{align}
& Z^0 = [e_\texttt{[CLASS]}, W_p v_1', \dots, W_p v_N'] + E_{pos} \\
& Z^i = \texttt{TransformerLayer}(Z_{i-1}; \theta_V^i)
\end{align}
where $Z^i = [z^i_0, \dots, z^i_N]$ is the hidden state of the layer $i$. The last hidden state of the \texttt{[CLASS]} embedding $z_0^l$ is selected to produce the representation of image $V$:
\begin{align}
\boldsymbol v = W_b z_0^l
\end{align}
where $\boldsymbol v$ is the image representation of $V$, and $W_b$ denotes a linear projection.

\noindent\textbf{Language Acquisition Encoder.}  As shown in the middle part of Figure \ref{fig: architecture}(a), the language acquisition encoder is built upon the pre-trained text encoder. Suppose $T$ is a sentence written in a non-native language $L$, we get the representation of $T$ through language acquisition encoder $\boldsymbol t = \Phi'(T; \theta_{\Phi}, \theta_{emb}, \theta_{L})$, where $\theta_{\Phi}$ are fixed parameters of the pre-trained text encoder, $\theta_{emb}$ refers to a shared non-native embedding block and $\theta_{L}$ represents specialized language acquirers for language $L$.
Non-native sentence $T$ is first tokenized and processed into word embeddings $E_T=[u_{0=\texttt{[SOS]}}, \dots, u_{M=\texttt{[EOS]}}]$ through the non-native embedding block. % We use the multilingual tokenizer of M-BERT \cite{bert} since it could handle 104 languages. 
% We also use the pre-trained weight of M-BERT to initialize the word embedding matrix as it speeds up convergence. 
The word embeddings are then encoded through the pre-trained transformer layers and language acquirers:
\begin{align}
X^0 &= [W_e u_{0=\texttt{[SOS]}}, W_e u_1, \dots, W_e u_{m=\texttt{[EOS]}}] + E_{pos} \\
H^i &= \texttt{TransformerLayer}(X^{i-1}; \theta_{\Phi}^i) \\ 
X^i &= \texttt{LA}(H^i; \theta_{L}^i)
\end{align}
where $X^i = [x^i_0, \dots, x^i_m]$ is the hidden state of the layer $i$. $W_e$ is a linear projection to keep dimension consistency. $\theta_{L}^i$ denotes the parameters of the $i$-th language acquirer for language $L$. Note that each non-native language has independent language acquirers, and all of them share the same word embedding block. As shown in Figure \ref{fig: architecture}(b), the language acquirer is implemented as a bottleneck MLP with residual connection \cite{resnet}:
\begin{align}
    & \texttt{LA}(X) =  W_{upper} \texttt{ReLU}(W_{down} X) + X
\end{align}

Similar with the pre-trained text encoder, the last hidden state of the \texttt{[EOS]} token is projected into the sentence representation $\boldsymbol t$:
\begin{align}
\label{eq: non-native_representation}
\boldsymbol t = W_a x^l_m
\end{align}
Note that Eq.\ref{eq: non-native_representation} shares the same linear projection $W_a$ with Eq.\ref{eq: native_representation}. 
The main advantage of the language acquisition encoder is that it can extend the VLP models to support new languages without influencing the existing languages, as it handles different languages with independent language acquirers.

\subsection{Training Strategy}
\label{subsec: training_strategy}
To simulate the language learning habits of humans, we optimize the model in two stages: the Native Language Transfer (NLT) stage and the Language Exposure (LE) stage.

%\paragraph
\noindent\textbf{Native Language Transfer.} 
When learning a new language, we humans initially tend to align it with the native language. To simulate this learning phase, we align the non-native representations to the native representations during the Native Language Transfer (NLT) stage. Specifically, suppose $\{(S_1, T_1), ..., (S_n, T_n)\}$ are translation pairs, where $S_i$ is in the native language, and $T_i$ is in a non-native language $L$. The objective in the NLT stage is minimizing the Mean Square Error (MSE) between the native representation $\boldsymbol s_i=\Phi(S_i; \theta_{\Phi})$ and the non-native representation $\boldsymbol t_i=\Phi'(T_i; \theta_{\Phi}, \theta_{L}, \theta_{emb})$:
\begin{equation}
    \mathcal{L}_{\mathrm{NLT}} = \frac{1}{B}\sum_{i=1}^{B}
    \Vert \boldsymbol s_i-\boldsymbol t_i \Vert ^2
\end{equation}
where B is the batch size. Note that $\theta_{\Phi}$ is loaded from the VLP model and is kept frozen. $\theta_{L}$ is trained for non-native language $L$. $\theta_{emb}$ is shared among non-native languages. 

During the NLT stage, the non-native correspondence with vision can be built pivoting on the native language, since the correspondence between the native language and vision is well established through VLP.

%\paragraph
\noindent\textbf{Language Exposure.}
After the NLT stage, the model has built an implicit connection between non-native languages and vision.  However, due to the existence of synonyms, two same words in the native language may correspond to different images. Thus, ambiguity may arise when learning non-native languages solely by relying on the native language. Actually, we can regard the language acquisition encoder after the NLT stage as a person with a certain language foundation. He/She has learned the basic usage of a language through native language teaching. To master it, 
he/she may practice the non-native language by interacting with the multimodal living environments. Inspired by this learning phase, we directly establish the cross-modal alignment between non-native languages and vision during the Language Exposure (LE) stage. Given image-text pairs $\{(V_1, T_1), ..., (V_n, T_n)\}$ where $T_i$ is in a non-native language $L$, the sentence representation $\boldsymbol t_i=\Phi^{'}(T_i; \theta_\Phi, \theta_{L}, \theta_{emb})$ should be closer to the aligned image representation $\boldsymbol v_i=\Psi(V_i; \theta_\Psi)$, and away from the misaligned one $\boldsymbol v_j=\Psi(V_j; \theta_\Psi),j\neq i$.
This can be achieved by performing contrastive learning between non-native languages and images. For a non-native sentence $T_i$, we treat the corresponding image $V_i$ as a positive sample, and other images in the same batch ${V_j}, j\neq i$ as negative samples. Vice versa for images. The objective in the LE stage is minimizing the NCE loss defined as follows:
\begin{align}
    & \mathcal{L}_{\mathrm{LE}} = \frac{1}{2}(\mathcal{L}_{v2t}+\mathcal{L}_{t2v}) \\
    & \mathcal{L}_{v2t} = -\frac{1}{B}\sum_{i=1}^{B}\mathrm{log}\frac{\mathrm{exp}(\mathrm{sim}(\boldsymbol v_i, \boldsymbol t_i)/\tau)}{\sum_{k=1}^N \mathrm{exp}(\mathrm{sim}(\boldsymbol v_i, \boldsymbol t_k)/\tau)} \\
    & \mathcal{L}_{t2v} = -\frac{1}{B}\sum_{i=1}^{B}\mathrm{log}\frac{\mathrm{exp}(\mathrm{sim}(\boldsymbol v_i, \boldsymbol t_i)/\tau)}{\sum_{k=1}^N \mathrm{exp}(\mathrm{sim}(\boldsymbol v_k, \boldsymbol t_i)/\tau)}
\end{align}
where $B$ is the batch size. $\mathrm{sim}(\boldsymbol x,\boldsymbol y)=\frac{\boldsymbol x^\top \boldsymbol y}{\norm{\boldsymbol x}\norm{\boldsymbol y}}$ is the cosine similarity between two vectors. $\tau$ is a temperature  hyper-parameter to scale the logits.
Note that though the image-to-text loss $\mathcal{L}_{v2t}$ is optimized, the pre-trained vision encoder is kept frozen during training. Similar to NLT, the trainable parameters in LE come from the language acquirers and the non-native embedding block.

\section{Experiments}
In this section, we first introduce the datasets used in this paper, and then present detailed experiments to evaluate the proposed MLA framework.
\subsection{Dataset Description}
We train our model with the Conceptual Captions (CC) dataset \cite{conceptual_caption} and two translation enhanced versions of the CC \cite{Zhou_2021_CVPR,multilingual-clip-github}. We use Multi30K \cite{m30k}, MSCOCO \cite{coco, coco-cn, coco-ja} and XTD \cite{xtd} for multilingual image-text retrieval evaluation, and MSRVTT \cite{Xu_2016_CVPR,mmp} for multilingual video-text retrieval evaluation. \\
%\subsubsection{Training Dataset}
\textbf{Conceptual Captions} (CC) \cite{conceptual_caption} contains 3.3 million image-text pairs in English crawled from the Web\footnote{We can only access $\sim$2.5 million images due to broken URLs.}. We also randomly select 300K image-text pairs denoted as \textbf{CC300K} for training our model to show the low-cost merit of MLA. For multilingual sentences, we leverage two translation augmented CC datasets: (1) \textbf{CC6L} \cite{Zhou_2021_CVPR} that translates all English captions of the CC into 5 languages (German(de), French(fr), Czech(cs), Chinese(zh)); and (2) \textbf{CC69L} \cite{multilingual-clip-github} that contains 27K captions in each of the 68 languages translated from English.\footnote{We remove captions of unaccessible images, leaving $\sim$20K captions for each language.} Considering the languages of the downstream datasets, we train the model with CC6L for multilingual image-text retrieval, and with CC69L for multilingual video-text retrieval. \\
%\subsubsection{Finetuning and Evaluation Dataset}
\textbf{Multi30K} \cite{m30k} is built upon Flickr30K \cite{young2014from}. The English(en) captions are manually translated into German(de), French(fr) and Czech(cs). It contains 31K images paired with 5 captions per image in English and German, and 1 caption in French and Czech. We use the standard train, dev and test splits defined in \cite{young2014from}. \\
\textbf{MSCOCO} \cite{coco} contains 123K images with 5 English captions per image. \cite{coco-ja} annotates 5 Japanese captions per image, and \cite{coco-cn} extends MSCOCO with Chinese captions for 20K images. We follow the standard train, dev and test splits for English and Japanese as in \cite{Karpathy_2015_CVPR}. For Chinese, we can only perform zero-shot evaluation on the test split defined in \cite{coco-cn}, as the full splits have overlaps with English and Japanese splits. \\
\textbf{XTD} \cite{xtd} provides captions in 11 languages (English(en), German(de), French(fr), Chinese(zh), Japanese(ja), Italian(it), Spanish(es), Russian(ru), Polish(pl), Turkish(tr), Korean(ko)) 
for 1K MSCOCO images. Except for Japanese, all non-English captions are translated from the English caption directly. We use this dataset for zero-shot image-text retrieval evaluation only. \\
\textbf{MSRVTT} \cite{Xu_2016_CVPR} is a video caption dataset with 10K videos, where each video is annotated with 20 English captions. Huang et al. translates the English captions into 8 languages (German(de), French(fr), Russian(ru), Spanish(es), Czech(cz), Swahili(sw), Chinese(zh) and Vietnamese(vi)) via machine translation service \cite{mmp}. We follow the standard train/dev splits in \cite{Xu_2016_CVPR}, and evaluate on the 1K test split as described in \cite{msrvtt-1k}.

\begin{table*}[htbp]
\centering
\scalebox{0.85}{
\begin{tabular}{lllcccccccc}
\toprule 
& \multirow{2}*{Method}& \multirow{2}*{Training Data} & \multicolumn{4}{c}{Multi30K} & \multicolumn{2}{c}{MSCOCO 1K} & \multicolumn{2}{c}{MSCOCO 5K} \\ \cmidrule(r){4-7} \cmidrule(r){8-9}
\cmidrule(r){10-11}
& &	&    en  &   de  &   fr  &   cs		&   en	& ja & en & ja \\
\midrule
\multirow{8}*{\rotatebox[]{90}{Zero-shot}} 
& Unicoder-VL &CC3M (English only)&	72.0&   - &   -&  -	&   63.7	& - & - & -\\
% & CLIP &WebImageText (English only)&	\underline{84.4}&   - &   -&  -	&   79.4	& - & 60.5 & -\\
% CLIP16		             		&   \textbf{86.4}&-& -&-&-&-&- \\
& ALIGN &AT-en (English only)&	84.3&	   - &-	&-	&\underline{80.0}	&-	& \underline{60.6} & -\\
& M$^3$P &CC3M+Wiki				&	57.9&   36.8&	27.1&  20.4	&   63.1	&   33.3 & - & -\\
& UC$^2$ &TrTrain(CC3M)			&	66.6&   62.5&	60.4&  55.1	&   70.9	&   62.3 & - & -\\
% & MKD$_{\mathrm{CLIP}}$ &TrTrain(CC+COCO+VizWiz)	&	81.0&   71.9&   71.1&  67.8	&   78.0	&   72.2	&   76.1 \\
% & MKD$_{\mathrm{CLIP}}$ &TrTrain(CC300k)&	82.1&   77.1&   75.2&  72.3	&   78.5	&   73.6	&   76.3 \\
& MURAL	&TrTrain(CC12M)+EOBT&	80.9&   76.0&   75.7&  68.2	&   78.1	&   72.5 & 58.0 & 49.7	\\
& MURAL$^{\dag}$ & AT+MBT &	82.4&   76.2&   75.0&  64.6	&   79.2	&   73.4 & 59.5 & \underline{54.4}	\\
& MLA$_{\mathrm{CLIP}}$	 &TrTrain(CC300K)& \underline{84.4}&   \underline{78.7}&   \underline{77.7}&  \underline{70.8}	&   79.4&   \underline{74.9} & 60.5 & 54.1\\
& MLA$_{\mathrm{CLIP16}}$ &TrTrain(CC300K)& \textbf{86.4}&   \textbf{80.8}&   \textbf{80.9}&   \textbf{72.9}&   \textbf{80.9}   &   \textbf{76.7} & \textbf{62.6} & \textbf{57.0} \\ \midrule
\multirow{4}*{\rotatebox[]{90}{FT-En}} 
& M$^3$P &CC3M+Wiki &	87.4&   82.1&	67.3&  65.0	&   88.6	&   56.0 & - & - \\
& UC$^2$ &TrTrain(CC3M) &	87.2&   \underline{83.8}&	77.6&  74.2	&   88.1	&   71.7 & - & -   \\
& MLA$_{\mathrm{CLIP}}$ &TrTrain(CC300K)&	\underline{92.0}&   82.6&   \underline{85.1}&  \underline{76.2}	&   \underline{89.3}	&   \underline{80.4} & \underline{75.7} & \underline{62.1} 	\\
& MLA$_{\mathrm{CLIP16}}$ &TrTrain(CC300K)&   \textbf{94.5}&   \textbf{86.4}&   \textbf{87.3}&   \textbf{79.5}&   \textbf{91.3}   &\textbf{82.6} & \textbf{79.4} & \textbf{65.5} \\ \midrule
\multirow{6}*{\rotatebox[]{90}{FT-All}} 
& M$^3$P$^\ddag$ &CC3M+Wiki &   87.7	&   82.7&   73.9&   72.2&   88.7$^\ddag$ & 87.9$^\ddag$ & - & - \\
& UC$^{2\ddag}$ &TrTrain(CC3M) &   88.2	&   84.5&   83.9&   81.2&   88.1$^\ddag$ & 87.5$^\ddag$ & - & - \\
& MURAL &TrTrain(CC12M)+EOBT & 91.0 &   87.3&   86.4&   82.4&   89.4&   87.4 & 73.7 & 71.9 \\
& MURAL$^{\dag}$ &AT+MBT&   \underline{92.2}	&   \underline{88.6}&   \underline{87.6}&   \underline{84.2}&   88.6&   \underline{88.4} & 75.4 & \underline{74.9} \\
& MLA$_{\mathrm{CLIP}}$ &TrTrain(CC300K) &   92.0	&   86.8&   85.4 & 82.3&   \underline{89.3} & 88.1 & \underline{75.7} & 73.2\\
& MLA$_{\mathrm{CLIP16}}$ &TrTrain(CC300K)&   \textbf{94.5}	&   \textbf{89.7}&   \textbf{89.2}&   \textbf{85.9}&   \textbf{91.3}&   \textbf{90.4} & \textbf{79.4} & \textbf{76.5} \\
\bottomrule
\end{tabular}
}
\vspace{-5pt}
\caption{Multilingual image-text retrieval results on Multi30K and MSCOCO. TrTrain: Translate-train, FT-En: \emph{Fine-tune on English}, FT-All: \emph{Fine-tune on All}. ${\dag}$: Models trained with publicly unavailable datasets. $\ddag$: Models fine-tuned on COCO-CN \cite{coco-cn}, which has an overlap train split with the test split of English and Japanese. Best results are in bold and second best are underlined.}
\label{tab: img-txt SOTA}
\vspace{-12pt}
\end{table*}

\begin{table}[t]
    \centering
    \scalebox{0.8}{
    \begin{tabular}{l|cl} \toprule
         Method & Trainable Params & Computing Costs \\ \midrule
         M$^3$P & 566 M & 4$\times$V100$\times$7d \\
         UC$^2$ & 478 M & 8$\times$V100$\times$4d \\
         MURAL & 300 M & 128$\times$TPUv3$\times$4d \\ \midrule
         Ours (MLA$_{\mathrm{CLIP}}$) & 108 M & 1$\times$V100$\times$0.5d \\ \bottomrule
    \end{tabular}
    }
    \vspace{-5pt}
    \caption{Comparison of trainable parameters and computing costs between MLA and M-VLPs.}
    \label{tab: resources}
    \vspace{-12pt}
\end{table}

\subsection{Implementation Details}
We apply MLA on two VLP models: CLIP-ViT-B-32 and CLIP-ViT-B-16 \cite{clip}, denoted as MLA$_{\mathrm{CLIP}}$ and MLA$_{\mathrm{CLIP16}}$ respectively. The hidden dimension of the language acquirers is set to 256, and all language acquirers for each non-native language cost only 3.14 MB parameters. The non-native embedding matrix is initialized with M-BERT \cite{bert}. It costs 92.2 MB and shared with all non-native languages. We train two separate models for multilingual image-text retrieval and video-text retrieval. For the image model, we train with CC6L \cite{Zhou_2021_CVPR}. For the video model, we use multilingual captions from CC69L \cite{multilingual-clip-github}. For both models, we optimize multiple language acquirers iteratively with a batch size of 128. The NLT stage performs 117,150 steps with a learning rate of 1e-4, and the LE stage performs 11,715 steps with a learning rate of 3e-6. The temperature $\tau$ is set to 0.01. For both stages, we use the Adam optimizer \cite{kingma2017adam} with a linear warm-up for the first 10\% of steps. The whole training process takes about 12 hours to converge on 1 Nvidia V100 GPU.

\begin{table*}[htbp]
    \centering
    \scalebox{0.8}{
    \begin{tabular}{clccccccccc|c}
        \toprule
        & Method & en & de & fr & cs & zh & ru & vi & sw & es & mean \\  \midrule
	   %\multicolumn{11}{l}{\emph{Zero-Shot}}\\	\midrule
	   \multirow{2}*{\rotatebox[]{90}{\small{ZS}}} 
        & Ours(MLA$_{\mathrm{CLIP}}$ w/o LE) & \textbf{30.8} & 18.3 & 18.9 & 14.5 & \textbf{18.6} &   12.6    &   7.2     &   10.2    &   19.3    &   16.7        \\
        & Ours(MLA$_{\mathrm{CLIP}}$)              &   \textbf{30.8}    &   \textbf{20.1}    &   \textbf{22.0}	&   \textbf{15.7}    &   18.3    &   \textbf{14.4}    &   \textbf{8.2}     &   \textbf{10.7}    &   \textbf{20.2}    &   \textbf{17.8}        \\  \midrule
	   %\multicolumn{11}{l}{\emph{Finetune on English}}	\\  \midrule
	   \multirow{2}*{\rotatebox[]{90}{\small{FT-En}}} 
        & XLM-R-MMP \cite{mmp}                &   23.8    &   19.4    &   20.7    &   19.3    &   18.2    &   19.1    &   8.2     &   8.4     &   20.4    &   17.5        \\  
        & Ours(MLA$_{\mathrm{CLIP}}$)              &   \textbf{42.5}    &   \textbf{26.1}    &   \textbf{26.7}    &   \textbf{20.5}    &   \textbf{25.3}    &   \textbf{18.9}    &   \textbf{12.9}    &   \textbf{12.6}    &   \textbf{27.2}    &   \textbf{23.6}        \\  \midrule
	   %\multicolumn{11}{l}{\emph{Finetune on All}}	\\  \midrule
	   \multirow{2}*{\rotatebox[]{90}{\small{FT-All}}} 
        & XLM-R-MMP \cite{mmp}                &   23.1    &   21.1    &   21.8    &   20.7    &   20.0    &   20.5    &   10.9    &   14.4    &   21.9    &   19.4        \\  
        % & Ours(MLA$_{\mathrm{CLIP}}$)              &   38.8    &   26.5    &   28.1    &   22.1    &   25.6    &   20.1    &   14      &   14.7    &   26.6    &   24.1        \\
        & Ours(MLA$_{\mathrm{CLIP}}$)     &   \textbf{42.5}    &   \textbf{33.1}    &   \textbf{34.5}    &   \textbf{30.5}    &   \textbf{31.6}    &   \textbf{28.9}    &   \textbf{16.9}    &   \textbf{24.3}    &   \textbf{33.5}    &   \textbf{30.6}        \\ 
        \bottomrule
    \end{tabular}
    }
    \vspace{-5pt}
    \caption{Multilingual video-text retrieval results on MSRVTT. ZS: \emph{Zero-shot}, FT-En: \emph{Fine-tune on English}, FT-All: \emph{Fine-tune on All}.}
    \vspace{-12pt}
    \label{tab:text_to_video_retrieval}
\end{table*}

\subsection{Evaluation on Multilingual Image-Text Retrieval}
\label{subsec: img-txt retrieval}
In multilingual image-text retrieval, models are given a sentence in a certain language to find the most semantically relevant image from an image database and vice versa. We compare our model with state-of-the-art multilingual vision-language pre-training methods under three settings:
\vspace{-12pt}
\begin{itemize}[leftmargin=*]
\setlength{\itemsep}{0pt}
\setlength{\parsep}{0pt} 
\setlength{\parskip}{0pt} 
\item \emph{\textbf{Zero-shot:}} we directly evaluate the model without fine-tuning on downstream datasets.
\item \emph{\textbf{Fine-tune on English:}} we first fine-tune the VLP model on downstream English data. We then insert the language acquirers and non-native embedding block into the fine-tuned model and evaluate on other languages directly.
\item \emph{\textbf{Fine-tune on All:}} after \emph{\textbf{Fine-tune on English}}, we fine-tune the language acquirers and non-native embedding block and freeze other parts of the model. 
\end{itemize}
\vspace{-10pt}
Following previous works \cite{m3p,Zhou_2021_CVPR,mural}, we report Average Recall (AR), which is the average score over Recall@1, Recall@5, and Recall@10 on two retrieval directions (image$\rightarrow$text, text$\rightarrow$image). The results are shown in Table \ref{tab: img-txt SOTA}. Also, the comparison of computing costs and parameters can be found in Table \ref{tab: resources}. 

Under the \textbf{\emph{Zero-shot}} setting, we observe that  MLA$_{\mathrm{CLIP}}$ performs significantly better than state-of-the-art M-VLP models on English. This is because MLA$_{\mathrm{CLIP}}$ could completely maintain the strong English performance of CLIP. In contrast, M-VLP models typically perform worse than their monolingual counterparts on English (M$^3$P 57.9 vs. Unicoder-VL \cite{li2020unicoder} 72.0, MURAL 80.9 vs. ALIGN \cite{align} 84.3). MLA$_{\mathrm{CLIP}}$ also outperforms M-VLP models on other languages. For example, MLA$_{\mathrm{CLIP}}$ achieves 78.7 average recall score on German, outperforming MURAL by 2.7\%. Note that the pre-training dataset of MURAL contains 12 million image-text pairs for each language, while MLA$_\mathrm{CLIP}$ only uses 300K training image-text pairs. It demonstrates that MLA is a high-data-efficient method to empower monolingual VLP models with multilingual capability. 
Under the \textbf{\emph{Fine-tune on English}} setting, MLA shows strong cross-lingual transfer capability. Under the \textbf{\emph{Fine-tune on All}} setting, MLA$_{\mathrm{CLIP}}$ performs slightly worse than MURAL which was pre-trained on publicly unavailable dataset AT+MBT \cite{mural}. We consider the reason is that MURAL has more trainable parameters than MLA$_{\mathrm{CLIP}}$ (300M vs 108M, as shown in Table \ref{tab: resources}) for fine-tuning, which makes it easier to fit the downstream datasets with a certain scale such as Multi30K and MSCOCO. 
% When we apply MLA on the CLIP ViT-B-16 (CLIP16),
MLA$_{\mathrm{CLIP16}}$ achieves state-of-the-art results on all languages under three settings. It indicates that if stronger VLP models such as ALIGN-L2 \cite{align} or Florence \cite{florence} are provided, better performance on multilingual image-text retrieval could be reached through MLA. 

\subsection{Evaluation on Multilingual Video-Text Retrieval}
In multilingual video-text retrieval, the model searches for the most semantically relevant videos given a text query in a certain language. Following \cite{luo2021clip4clip}, we first uniformly sample 12 frames from each video, and use the pre-trained vision encoder to extract representations for each frame. We then perform mean pooling over frame representations to get the video representation. 

We also evaluate the models under three settings as in Sec.\ref{subsec: img-txt retrieval}.
We report the text$\rightarrow$video Recall@1 score in Table \ref{tab:text_to_video_retrieval}.
Under \textbf{\textit{Zero-shot}} setting, MLA$_\mathrm{CLIP}$, which is trained on CC69L without using any video data, achieves comparable or even better results than the fine-tuning results of the state-of-the-art M-VLP model XLM-R-MMP \cite{mmp} on several languages (de: 20.1 vs. 21.1; fr: 22.0 vs. 21.8; es: 20.2 vs. 21.9). Under the \textbf{\textit{Fine-tune on English}} and \textbf{\textit{Fine-tune on All}} settings, MLA$_\mathrm{CLIP}$ also outperforms XLM-R-MMP significantly. We consider the convincing performance comes from two reasons: 1) CLIP is a strong VLP model that can generalize well on video data. 2) The proposed MLA framework can well transfer the open-domain knowledge learned by CLIP to other languages. These results suggest that MLA could maintain the open-domain capability of the VLP model which generalizes well on different downstream data.

\subsection{Ablation Study}
\noindent\textbf{\textit{{A. Training Strategy}}}

We conduct an ablation study in Table \ref{tab: ablation_training_strategy} to validate the effectiveness of the proposed MLA training strategy. For those settings with NLT and LE at the same stage, we add the loss of the two objectives together during training. By comparing row 1 to row 2\&3, we observe that LE at stage one leads to poor performance. This indicates that aligning with the native language is more important for the VLP model to acquire new languages at an early stage. It is consistent with the learning habits of humans. By comparing row 1 and row 4, we see that LE at stage two could bring improvements on the new languages. Additionally, comparing row 4 and row 5 suggests that optimizing the model with NLT and LE together at stage two does not bring improvements.
\begin{table}[ht]
\centering
\resizebox{\linewidth}{!}{
    \begin{tabular}{c|ccccccccc}
    \toprule
    \multirow{2}{*}{Row} & \multicolumn{2}{c}{Stage one} & \multicolumn{2}{c}{Stage two} & \multicolumn{3}{c}{Multi30K} & \multicolumn{2}{c}{MSCOCO 1K} \\ \cmidrule(r){6-8} \cmidrule(r){9-10} 
    & NLT & LE & NLT & LE & de & fr & cs & ja & zh \\ \midrule
    % 1* & nce & & & & 63.0 & 58.5 & 49.6 & 57.1 & 64.8 \\
    % 2* & & mse & & & 47.2 & 47.0 & 37.4 & 47.5 & 54.9 \\
    1 &	\checkmark & & & & 76.3 & 74.2 & 67.2 & 72.1 & 75.7 \\ 
    2 &	& \checkmark & & & 68.2 & 67.7 & 58.6 & 65.9 & 71.7 \\ 
    3 &	\checkmark & \checkmark & & & 71.1 & 69.7 & 59.8 & 67.6 & 73.9    \\
    4 &	\checkmark & & & \checkmark & \textbf{78.7} & \textbf{77.7} & \textbf{70.8} & \textbf{74.9} & \textbf{78.5} \\   
    5 & \checkmark & & \checkmark & \checkmark & 78.4 & 77.3 & 69.9 & 74.2 & 78.1 \\ \bottomrule 
    \end{tabular}
}
\vspace{-5pt}
\caption{Ablation study on training strategy}
\label{tab: ablation_training_strategy}
\vspace{-12pt}
\end{table}

\noindent\textbf{\textit{{B. Language Acquirers and Embedding Initialization}}}

In order to validate the effectiveness of the proposed Language Acquirers, we remove the language acquirers and the M-BERT embedding initialization from the model respectively and evaluate on zero-shot multilingual image-text retrieval. As shown in Table \ref{tab: LA_vs_EI}, the performance on all languages drops significantly without language acquirers. Meanwhile, initializing the embedding with M-BERT \cite{bert} only brings incremental improvements. It indicates that the language acquirers contribute most to the performance, and MLA does not depend much on the initialization of non-native embedding.

\begin{table}[!h]
\centering
\scalebox{0.8}{
    \begin{tabular}{l|ccccc}
        \toprule
        % Methods & de & fr & cs & ja & zh \\ \midrule
        \multirow{2}*{Methods} & \multicolumn{3}{c}{Multi30K} & \multicolumn{2}{|c}{MSCOCO 1K} \\
         & de & fr & cs & \multicolumn{1}{|c}{ja} & zh \\ \midrule
        MLA$_{\mathrm{CLIP}}$ & \textbf{78.7} & \textbf{77.7} & \textbf{70.8} & \multicolumn{1}{|c}{\textbf{74.9}} & \textbf{78.5} \\ % \midrule
        MLA$_{\mathrm{CLIP}}$ w/o LA & 76.1 & 74.9 & 65.7 & \multicolumn{1}{|c}{70.3} & 76.5 \\
        MLA$_{\mathrm{CLIP}}$ w/o EI & 77.9 & 76.2 & 69.4 & \multicolumn{1}{|c}{74.6} & 78.1 \\
        \bottomrule
    \end{tabular}
    }
    \vspace{-5pt}
    \caption{Ablation study on language acquirers and embedding initialization. LA: Language Acquirers, EI: M-BERT Embedding Initialization}
    \label{tab: LA_vs_EI}
    \vspace{-8pt}
\end{table}

\begin{table*}[ht]
    \centering
    \scalebox{0.8}{
    \begin{tabular}{c|l|ccccccccccc}
        \toprule
        
        \multirow{2}{*}{Row} & \multirow{2}*{Method} & \multicolumn{5}{c}{Seen languages} & \multicolumn{6}{|c}{Unseen languages} \\ 
        & & en & de & fr & zh & ja & \multicolumn{1}{|c}{it} & es & ru & pl & tr & ko \\  \midrule
        1 & UC$^2$ w/o unseen language training & 71.8 & 67.5 & 68.4 & 61.9 & 51.5 & \multicolumn{1}{|c}{-}	& -	& -	& - & - & \\
        2 & UC$^2$ w/ unseen language training & 63.6 & 57.8 & 57.6 & 57.6 & 48.4 & \multicolumn{1}{|c}{56.4} & 56.2 & 51.3 & 56.4 & 51.62 & 51.3 \\
        3 & UC$^2$ w/ all language training & 65.2 & 59.3 & 59.7 & 60.1 & 50.5 & \multicolumn{1}{|c}{57.7} & 56.5 & 50.9 & 55.3 & 53.2 & 50.2 \\ \midrule		
		4 & MLA$_{\mathrm{CLIP}}$ w/o unseen language training & 75.9 & \textbf{72.6} & \textbf{72.9} & 73.7 & \textbf{67.2} & \multicolumn{1}{|c}{-} & - & - & - & - & - \\
		5 & MLA$_{\mathrm{CLIP}}$ w/ unseen language training & \textbf{76.0} & \textbf{72.6} & \textbf{72.9} & \textbf{73.8} & \textbf{67.2} & \multicolumn{1}{|c}{\textbf{64.7}} & \textbf{62.8} & \textbf{58.1} & \textbf{63.0} & \textbf{56.5} & \textbf{57.3} \\ \bottomrule
    \end{tabular}
    }
    \vspace{-5pt}
    \caption{Language extention experiments on XTD dataset.}
    \label{tab:xtd}
    \vspace{-12pt}
\end{table*}

%\subsubsection{Multilingual Acquisition vs. Cross-modal Acquisition}

\noindent\textbf{\textit{{C. Low-resource Languages}}}

Image-text pairs may be rare for low-resource languages. 
To explore the performance of MLA under this situation, we further simulate a \textbf{low-resource scenario} using XTD dataset. We finetune MLA$_\mathrm{CLIP}$ and UC$^2$ (pre-trained on CC6L) with small amount of data from XTD in an unseen language. We randomly sample 600 pairs for finetuning, and the remained 400 samples are evenly divided for validation and testing. Korean is chosen to perform simulation as its script and language family are not covered by CC6L. Experimental results in Table \ref{tab: low resource} show that MLA can achieve competitive results with \textbf{very small amount of text-text pairs only} (row 2), and adding image-text pairs brings further improvement (row 3). It demonstrates that MLA is still an attractive method for low-resource languages even without any image-text pairs.
\begin{table}[htpb]
\centering
\vspace{-0pt}
    \scalebox{0.8}{
    \vspace{-12pt}
    \begin{tabular}{l|ll|cc}
    \hline
    & \multirow{2}*{Methods} & \multirow{2}*{Data} & \multicolumn{1}{c}{Training samples} \\ % \cline{4-5}
    & & & \multicolumn{1}{c}{100 / 200 / 600} \\ \hline
    1 & UC$^2$ & Img-Txt & 47.0 / 60.1 / 78.3 \\
    2 & MLA$_\mathrm{CLIP}$ & Txt-Txt & 51.7 / 62.8 / 78.7 \\
    3 & MLA$_\mathrm{CLIP}$ & Both & \textbf{56.7} / \textbf{66.9} / \textbf{80.1} \\ \hline
    \end{tabular}
    }
    \caption{Low resource performance on image-Korean retrieval.}
    \label{tab: low resource}
    \vspace{-12pt}
\end{table}

%\subsection{Analyze}

%\subsubsection{Size of Training Corpora}
\noindent\textbf{\textit{{D. Amount of Training Data}}}

\begin{figure}[!ht]
    \centering
    \includegraphics[width=\linewidth]{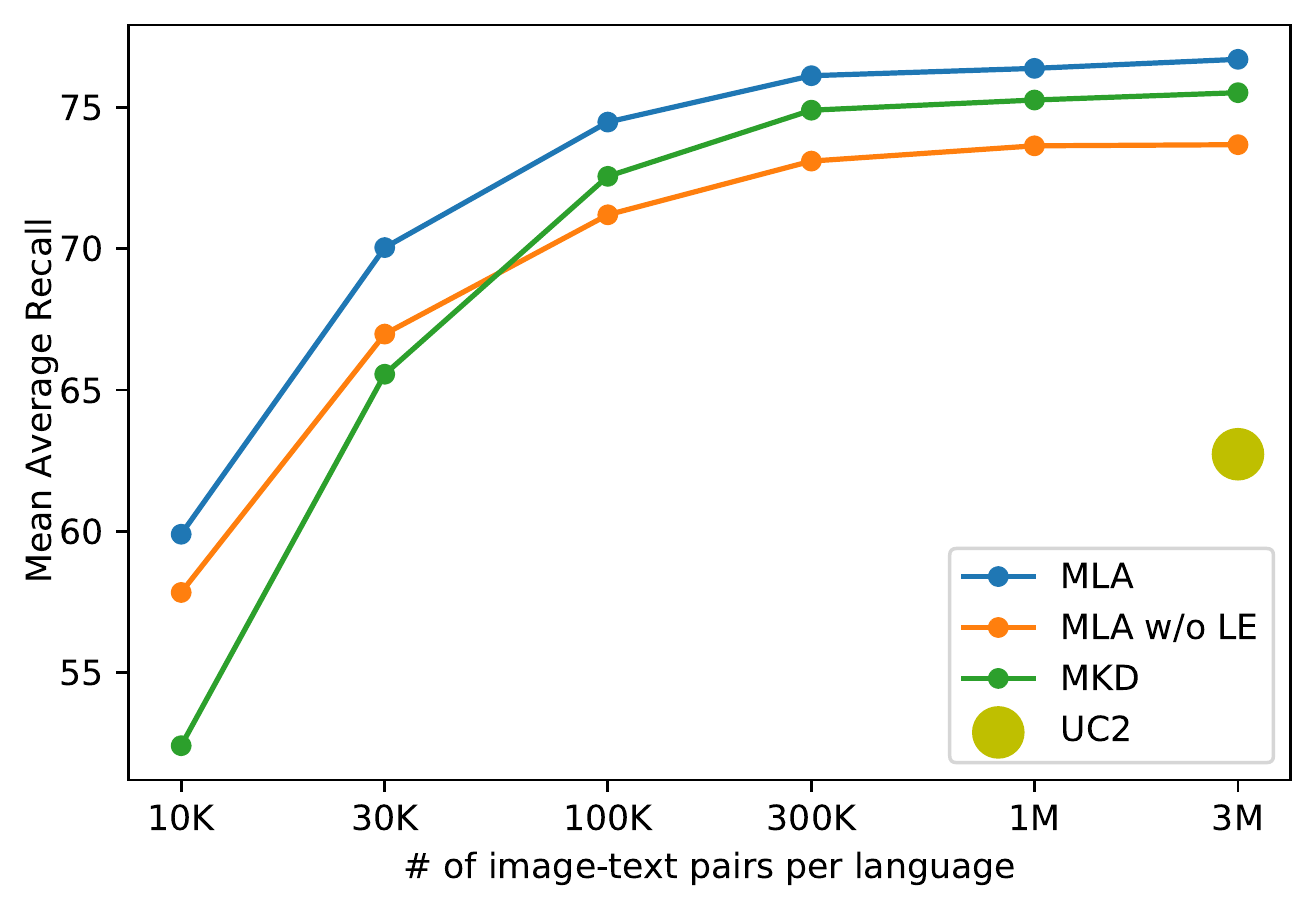}
    \vspace{-18pt}
    \caption{Mean AR vs. number of image-text pairs per language.}
    \label{fig:training_nums}
    \vspace{-12pt}
\end{figure}

Multilingual image-text pairs may be rare in practice. 
% It is expensive to get multilingual image-text pairs for building multilingual multimodal models. 
To explore the performance of MLA under low-resource conditions, we conduct experiments to control the numbers of image-text pairs used for each language. We train the models with CC6L and evaluate on MSCOCO 1K and Multi30K under the zero-shot setting. The corresponding mean AR over non-English languages (de, fr, cs, ja, zh) are drawn in Figure \ref{fig:training_nums}. We observe that MLA performs significantly better than MKD \cite{mse-kd} in all cases. Note that when the amount of training data is small, the advantage of MLA is more obvious, which could outperform MKD even without the LE training stage. 
Additionally, when training with only 30K image-text pairs per language, MLA outperforms UC$^2$, which is pre-trained with 3M pairs per language. MLA is thus a data-efficient method to build multilingual and multimodal models.

\newpage
\noindent\textbf{\textit{{E. Language Extensibility}}}

Multilingual models often encounter the need to support new languages that do not occur in the training stage. 
We conduct language extension experiments to compare MLA$_\mathrm{CLIP}$ with M-VLP model UC$^2$ \cite{Zhou_2021_CVPR} on the XTD dataset \cite{xtd}. XTD supports 11 languages, and 5 of them (en, de, fr, cs, zh, ja) are seen in the pre-training stage of UC$^2$, while other 6 languages (it, es, ru, pl, tr, ko) are unseen. To make a fair comparison, we first train MLA$_\mathrm{CLIP}$ with the same data as UC$^2$ and then train both of them on unseen languages with CC69L. The zero-shot image-text retrieval results on XTD are shown in Table \ref{tab:xtd}. We observe a significant performance degeneration on the seen languages for UC$^2$ when training solely with unseen languages (row 1 vs. row 2). Even keep training with the seen languages, the performance is still significantly reduced due to the limited model capacity (row 1 vs. row 3). In contrast, as MLA decoupled multiple languages through acquirers, the performance of the seen languages is rarely affected (row 4 vs. row 5) . This suggests that MLA framework can build multimodal multilingual models that are suitable for supporting increasing numbers of languages. 

\section{Conclusion}
In this paper, we propose the MultiLingual Acquisition (MLA) framework that can generalize monolingual Vision-Language Pre-training models into multilingual with low-cost and high-flexibility.
MLA injects language acquirers and a non-native embedding block into VLPs to support new languages. Inspired by the language learning habits of humans, we propose a two-stage training strategy to optimize the language acquirers and non-native embedding block. MLA applied on CLIP achieves state-of-the-art performances on multilingual image-text and video-text retrieval benchmarks with much less computing costs and training data. 
Extensive ablation studies demonstrate that MLA is a flexible, effective, and efficient method to build multimodal and multilingual models. 

\bibliography{bib}
\bibliographystyle{icml2022}

%%%%%%%%%%%%%%%%%%%%%%%%%%%%%%%%%%%%%%%%%%%%%%%%%%%%%%%%%%%%%%%%%%%%%%%%%%%%%%%
%%%%%%%%%%%%%%%%%%%%%%%%%%%%%%%%%%%%%%%%%%%%%%%%%%%%%%%%%%%%%%%%%%%%%%%%%%%%%%%
% APPENDIX
%%%%%%%%%%%%%%%%%%%%%%%%%%%%%%%%%%%%%%%%%%%%%%%%%%%%%%%%%%%%%%%%%%%%%%%%%%%%%%%
%%%%%%%%%%%%%%%%%%%%%%%%%%%%%%%%%%%%%%%%%%%%%%%%%%%%%%%%%%%%%%%%%%%%%%%%%%%%%%%
\newpage
\appendix
\onecolumn

\appendix
\section{Qualitative Analysis}
\subsection{Case study}
\begin{figure*}[htbp]
    \centering
    \includegraphics[width=\linewidth]{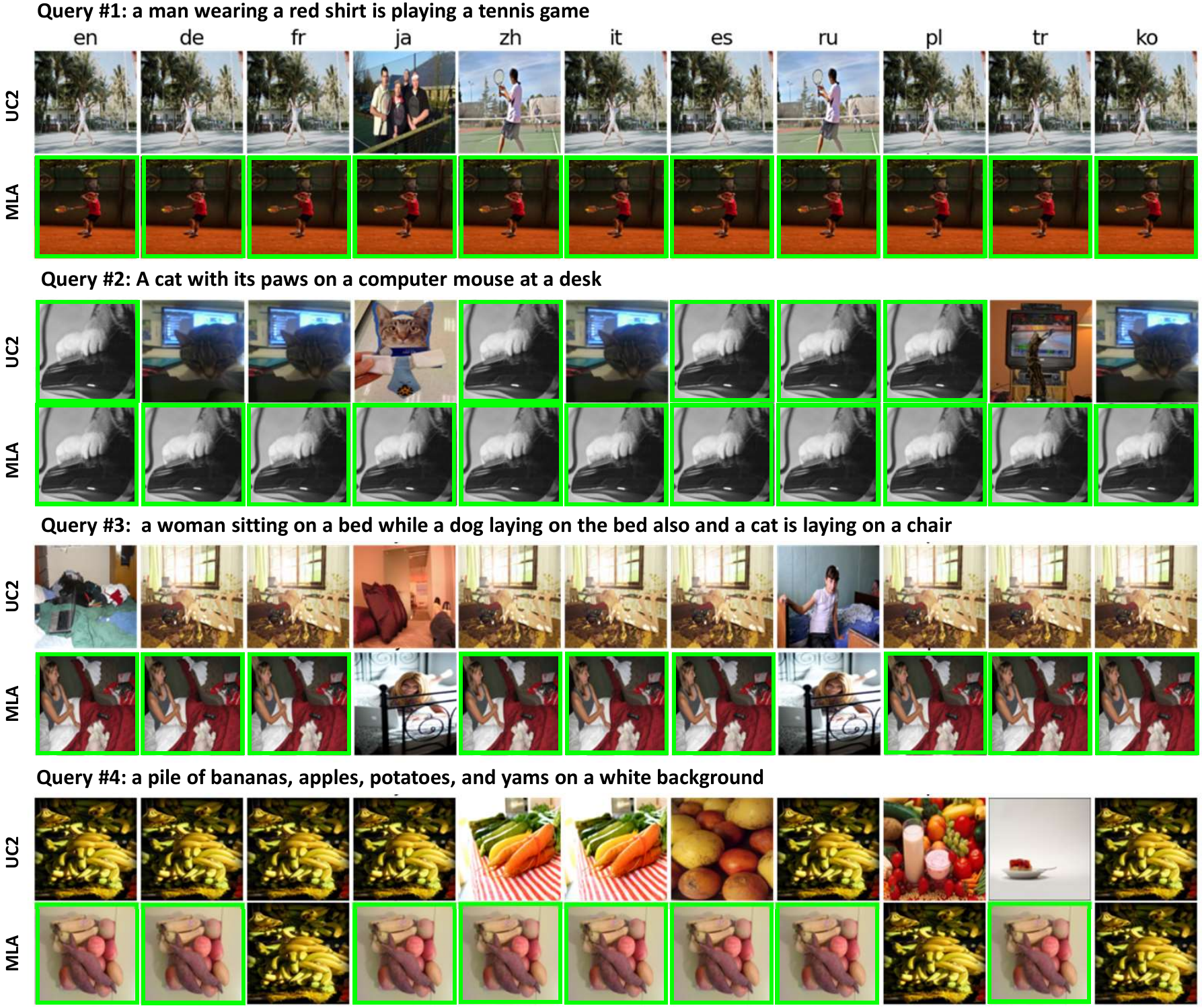}
    \caption{Top-1 retrieved images for given text queries in 11 languages on XTD dataset. Only English queries are shown in this figure. The correct images are bordered green.}
    \label{fig: case study}
\end{figure*}

In Figure \ref{fig: case study}, we visualize the top-1 retrieved images for given text queries in 11 languages on XTD dataset \cite{xtd}. Compared with the multilingual vision-language pre-training model UC$^2$ \cite{Zhou_2021_CVPR}, MLA can better capture entities, attributes, and actions to retrieve the correct image. Specifically, given simple queries that contain few entities such as Query \#1 or Query \#2, the images retrieved by MLA show high consistency across languages, since the representations of non-English queries are aligned to English in the NLT stage. For the more complex queries such as Query \#3 or Query \#4, MLA also shows better fidelity to all entities in most cases.

\newpage

\subsection{Representation visualization}
\begin{figure*}[htbp]
    \centering
    \includegraphics[width=\linewidth]{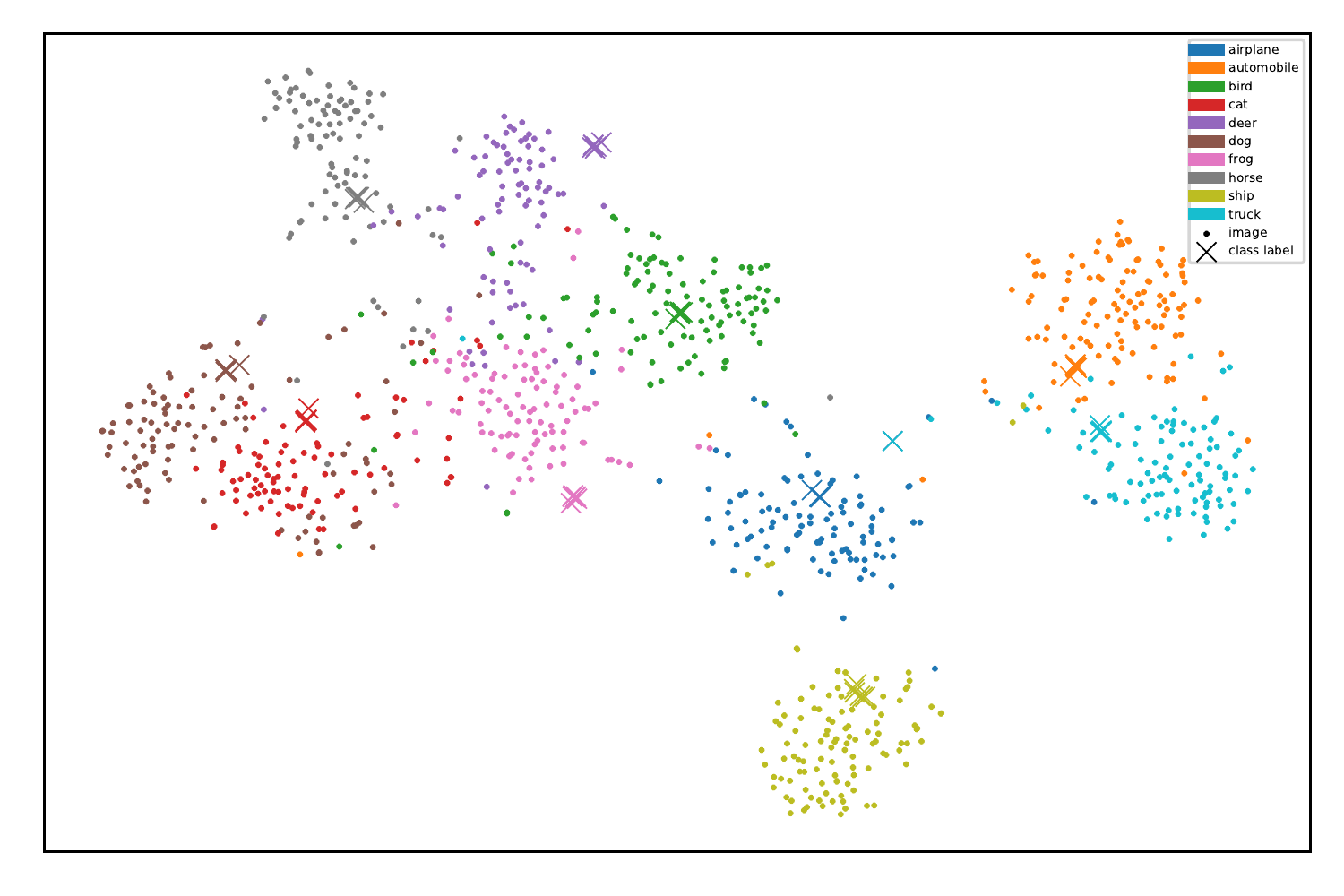}
    \caption{Representation visualization with t-SNE. The categories are color coded. '$\tiny {\bullet}$' denotes a image representation, and '$\times$' denotes a class label representation in a certain language. }
    \label{fig: representation visiualize}
\end{figure*}

To visualize the multimodal and multilingual representation space, we translate the English class labels of CIFAR10 \cite{cifar} into 5 languages including German (de), French (fr), Czech (cs), Chinese (zh), and Japanese (ja). The images and labels in 6 languages are encoded into representations through MLA$_\mathrm{CLIP}$. Figure \ref{fig: representation visiualize} shows the t-SNE \cite{van2008visualizing} visualization of these representations. We can see that the representations from different languages and modalities are clustered according to the semantics. It suggests that MLA$_\mathrm{CLIP}$ indeed can project images and multilingual sentences into a shared multimodal and multilingual space.
%\newpage
\section{Additional Ablation studies}
We conduct additional ablation studies to verify the effectiveness of MLA. All experiments in this section are conducted on zero-shot image-text retrieval.

\subsection{Structure of language acquirer}
In our proposed MLA, we implement the language acquirer as a bottleneck MLP. In Table.\ref{tab: mlp}, we compare the different structure of the language acquirer, the bottleneck MLP and a linear projection layer with the same amount of parameters. MLP works slightly better than the linear projection. Thus, we choose MLP to conduct our major experiments.  

\begin{table}[ht]
\centering
\caption{Ablation study on structure of language acquirer.}
\begin{tabular}{l|c|ccccc}
    \hline
    \multirow{2}*{Methods} & \multirow{2}*{Component} & \multicolumn{3}{c}{Multi30K} & \multicolumn{2}{|c}{MSCOCO 1K} \\
     & & de & fr & cs & \multicolumn{1}{|c}{ja} & zh \\ \hline
    MLA$_{\mathrm{CLIP}}$ & Linear & 78.2 & 77.6 & 69.3 & \multicolumn{1}{|c}{74.6} & 78.0 \\ 
    MLA$_{\mathrm{CLIP}}$ & MLP & \textbf{78.7} & \textbf{77.7} & \textbf{70.8} & \multicolumn{1}{|c}{\textbf{74.9}} & \textbf{78.5} \\
    \hline
\end{tabular}
\label{tab: mlp}
\vspace{-18pt}
\end{table}

\subsection{Objectives in the two-stage training}
In the default setting, we use the MSE objective during the NLT stage and the NCE objective during the LE stage. The MSE objective requires paired representations to be completely consistent, while the NCE objective only requires positive pairs to be closer than negative ones. We conduct experiments to use different objectives in the two stages. As shown in Table \ref{tab: loss_ablation}, we observe that the MSE objective is more suitable for the NLT (row 1 vs. row 2, row 7 vs. row 8) stage, and the NCE objective performs better for the LE stage (row 3 vs. row 4, row 5 vs. row 6). We consider the reason is that in the NLT stage, we leverage translation pairs to build alignment between languages. Since the two sentences of a translation pair are highly semantically related, their representations can be very similar. Thus, optimizing a strong objective like MSE during the NLT stage is feasible. However, during the LE stage, the optimization is conducted with image-text pairs. Although the image and text are semantically related, one sentence can hardly describe all the information in the image. Therefore, a weak objective like NCE is suitable for the LE stage.
\begin{table}[htpb]
\centering
\caption{Ablation study on objectives in the two training stages. mse: MSE objective, nce: NCE objective}
\scalebox{1.0}{
    \begin{tabular}{c|ccccccccc}
    \toprule
    \multirow{2}{*}{Row} & \multicolumn{2}{c}{Stage one} & \multicolumn{2}{c}{Stage two} & \multicolumn{3}{c}{Multi30K} & \multicolumn{2}{c}{MSCOCO 1K} \\ \cmidrule(r){6-8} \cmidrule(r){9-10} 
    & NLT & LE & NLT & LE & de & fr & cs & ja & zh \\ \midrule
    1 & mse & & & & 76.3 & 74.2 & 67.2 & 72.1 & 75.7 \\
    2 & nce & & & & 63.0 & 58.5 & 49.6 & 57.6 & 64.8 \\
    3 & & mse & & & 47.2 & 47.0 & 37.4 & 46.3 & 54.9 \\
    4 & & nce & & & 68.2 & 67.7 & 58.6 & 65.9 & 71.7 \\
    5 & mse & & & mse & 55.0 & 51.3 & 43.8 & 50.9 & 57.9 \\
    6 & mse & & & nce & \textbf{78.7} & \textbf{77.7} & \textbf{70.8} & \textbf{74.9} & \textbf{78.5} \\
    7 & mse & & mse & nce & 78.4 & 77.3 & 69.9 & 74.2 & 78.1 \\ 
    8 & mse & & nce & nce & 78.1 & 77.2 & 69.5 & 73.9 & 78.2 \\ 
    \bottomrule
    \end{tabular}
}
\label{tab: loss_ablation}
\end{table}

\subsection{Multilingual Acquisition vs. Cross-modal Acquisition}
MLA adopts the "Multimodal$\rightarrow$Multilingual" strategy that empowers VLP models with multilingual capability. However, there is another option of "Multilingual$\rightarrow$Multimodal" that empowers multilingual pre-training models with multimodal capability. To make a comparison between these two strategies, we implement the Cross-Modal Acquisition (CMA) that inserts cross-modal acquirers in each layer of the multilingual pre-training model M-BERT \cite{bert}. 
We keep the pre-trained M-BERT fixed and train the cross-modal acquirers with the same two-stage strategy as MLA. From Table \ref{tab: LA_vs_MA}, we find that CMA performs worse than MLA in all languages. It suggests that generalizing multilingual models to multimodal is harder than generalizing multimodal models to multilingual through lightweight acquirers.  
% \haw{not necessary:
% Additionally, as CMA does not use the pre-trained text encoder of VLP during inference, it can not maintain the performance on English. Thus, a "Multimodal$\rightarrow$Multilingual" strategy like MLA is a better choice to achieve multimodal and multilingual capability.}

\begin{table}[htbp]  % {r}{0.5\linewidth}
\centering
\caption{Multilingual Acquisition vs. Cross-modal Acquisition}
% \scalebox{0.8}{
    \begin{tabular}{l|ccccccc}
        \toprule
        \multirow{2}*{Methods} & \multicolumn{4}{c}{Multi30K} & \multicolumn{3}{c}{MSCOCO 1K} \\
        & en & de & fr & cs & en & ja & zh	\\  \midrule
        CMA$_{\mathrm{CLIP}}$ & 80.2 & 73.9 & 72.8 & 67.0 & 76.3 & 69.8 & 75.1 \\
%	   MKD$_{\mathrm{CLIP}}$ & 82.1 & 77.1 & 75.2 & \textbf{72.3} & 78.5 & 73.6 & 76.3 \\
        MLA$_{\mathrm{CLIP}}$ & \textbf{84.4} & \textbf{78.7} & \textbf{77.7} & \textbf{70.8} & \textbf{79.4} & \textbf{74.9} & \textbf{78.5} \\
        \bottomrule
    \end{tabular}
% }
    \label{tab: LA_vs_MA}
\end{table}

\section{Open-domain Image Classification}
%Though MKD \cite{mse-kd} could reach comparable results on image-text retrieval with MLA, 
In order to test the open-domain capability of models, we conduct zero-shot open-domain image classification experiments on CIFAR100 \cite{cifar}, ImageNet-V2 \cite{imagenetv2}, ImageNet-R \cite{imagenet-r} and ImageNet-A \cite{imagenet-a} datasets. 
As shown in Table \ref{tab: image_classification}, MKD \cite{mse-kd} performs badly on open-domain image classification. We consider the reason is that MKD abandons the original text encoder which contains open-domain multimodal knowledge from large-scale pre-training. In contrast, MLA keeps the original text encoder fixed and thus could maintain the open-domain capability of the pre-training model. % so it could perform well under open-domain scenarios.

\begin{table}[htbp]
\centering
\caption{Top-1 Accuracy of zero-shot open-domain image classification.}
\scalebox{1.0}{
	\begin{tabular}{lcccc}
	\toprule
	Methods			& CIFAR100 & ImageNet-V2 & ImageNet-R & ImageNet-A \\ \midrule
	MKD$_{\mathrm{CLIP}}$		& 32.8     & 54.7       	&  37.7      &  23.5       \\
	MLA$_{\mathrm{CLIP}}$		& \textbf{64.2}     & \textbf{63.4}        & \textbf{69.0}       & \textbf{31.4}      \\ \bottomrule
	\end{tabular}
	}
	\label{tab: image_classification}
\end{table}

% You can have as much text here as you want. The main body must be at most $8$ pages long.
% For the final version, one more page can be added.
% If you want, you can use an appendix like this one, even using the one-column format.
%%%%%%%%%%%%%%%%%%%%%%%%%%%%%%%%%%%%%%%%%%%%%%%%%%%%%%%%%%%%%%%%%%%%%%%%%%%%%%%
%%%%%%%%%%%%%%%%%%%%%%%%%%%%%%%%%%%%%%%%%%%%%%%%%%%%%%%%%%%%%%%%%%%%%%%%%%%%%%%

\end{document}